\pdfoutput=1

\documentclass[11pt]{article}

\usepackage[]{naacl2021}

\usepackage{times}
\usepackage{latexsym}
\usepackage{algpseudocode}
\usepackage{algorithm}
\usepackage{graphicx}
\usepackage{hyperref}

\usepackage[T1]{fontenc}

\usepackage[utf8]{inputenc}

\usepackage{microtype}

\title{GateNLP-UShef at SemEval-2022 Task 8: Entity-Enriched Siamese Transformer for Multilingual News Article Similarity}

\author{Iknoor Singh, Yue Li, Melissa Thong {\normalfont and} Carolina Scarton \\
  Department of Computer Science, University of Sheffield (UK)\\
  \texttt{\{i.singh, yli381, mlthong1, c.scarton\}@sheffield.ac.uk} \\}

\begin{document}
\maketitle

\begin{abstract}
This paper describes the second-placed system on the leaderboard of SemEval-2022 Task 8: Multilingual News Article Similarity. We propose an entity-enriched Siamese Transformer which computes news article similarity based on different sub-dimensions, such as the shared narrative, entities, location and time of the event discussed in the news article. Our system exploits a Siamese network architecture using a Transformer encoder to learn document-level representations for the purpose of capturing the narrative together with the auxiliary entity-based features extracted from the news articles. The intuition behind using all these features together is to capture the similarity between news articles at different granularity levels and to assess the extent to which different news outlets write about ``the same events''. Our experimental results and detailed ablation study demonstrate the effectiveness and the validity of our proposed method. 
\end{abstract}

\section{Introduction}
News article similarity measures could facilitate various important tasks such as the clustering of news \cite{montalvo2007bilingual, azzopardi2012incremental}, duplicate news detection \cite{alonso2013duplicate,gibson2008identification, singh2021false, theobald2008spotsigs}, fact-checking \cite{hassan2017claimbuster, jiang2021categorising} and tracking of the spread of news \cite{zhai2005tracking}. SemEval-2022 task 8 \citep{taskpaper} assesses the similarity between news articles in terms of the real world happenings. Therefore, it mainly considers the location, time, entities and narratives, instead of writing style, political spin, or tone of the article. The training set comprises of monolingual and cross-lingual news articles pairs in 10 different languages. 

The main contributions in this paper are: 1) We propose an entity-enriched Siamese Transformer whose general idea is to enable the model to explicitly learn from entity features (geolocation, organization, date and quantity) that are crucial to determine the similarity between news events but difficult to extract directly from the language models during fine-tuning. 2) We explore different data augmentation approaches through semi-supervised learning in order to tackle the imbalanced data problem. 3) We compare our proposed model with strong baselines and conduct an ablation study to analyse the contribution of each component in our model. We also present an error analysis at the end of this paper. Our best system which exploits Language Agnostic BERT Sentence Representations \citep{feng2020language} ranks 2nd in the competition.

\section{Background}
The goal of the task is to predict similarity scores ranging from 1 (most similar) to 4 (least similar) for a give news article pair. The training data 
consists of 4,964 article pairs in seven distinct languages (English, German, Spanish, Turkish, Polish, Arabic and French), with seven groups of monolingual pairs and only one group of cross-lingual pair (German and English). The 4,902 pairs in test data feature three more languages (Chinese, Russian and Italian) and seven groups of new cross-lingual pairs. Appendix \ref{appendix:datadist} (Figure \ref{fig:data_distri}) shows the distribution of the train and test data. Most of the news pairs exhibit low level of similarity. However, the test set contains more pairs with the same news stories. The similarity based on location, time, entities, narratives, style and tone is also annotated respectively and given for the training data. 


Previous studies on textual similarity have investigated different approaches based on corpus, knowledge or deep neural network \cite{chandrasekaran2021evolution,gomaa2013survey,devlin2018bert,reimers2019sentence,thakur2020augmented, singh2021multistage}, but they mainly experiment with short text pairs (e.g., the Semantic Textual Similarity (STS) benchmark \cite{cer2017semeval}, which is a sentence-level task). Measurement of the document similarity is arguably more challenging than that of short text since the information in the document is sparse and the model is easier to be misled by non-essential content. This paper tries to tackle the above mentioned challenges to build robust models for document similarity (Section \ref{sys}).

\section{System Overview}
\label{sys}

\begin{figure*}[!htbp]
    \centering
    \includegraphics[width=14cm,height=5.5cm]{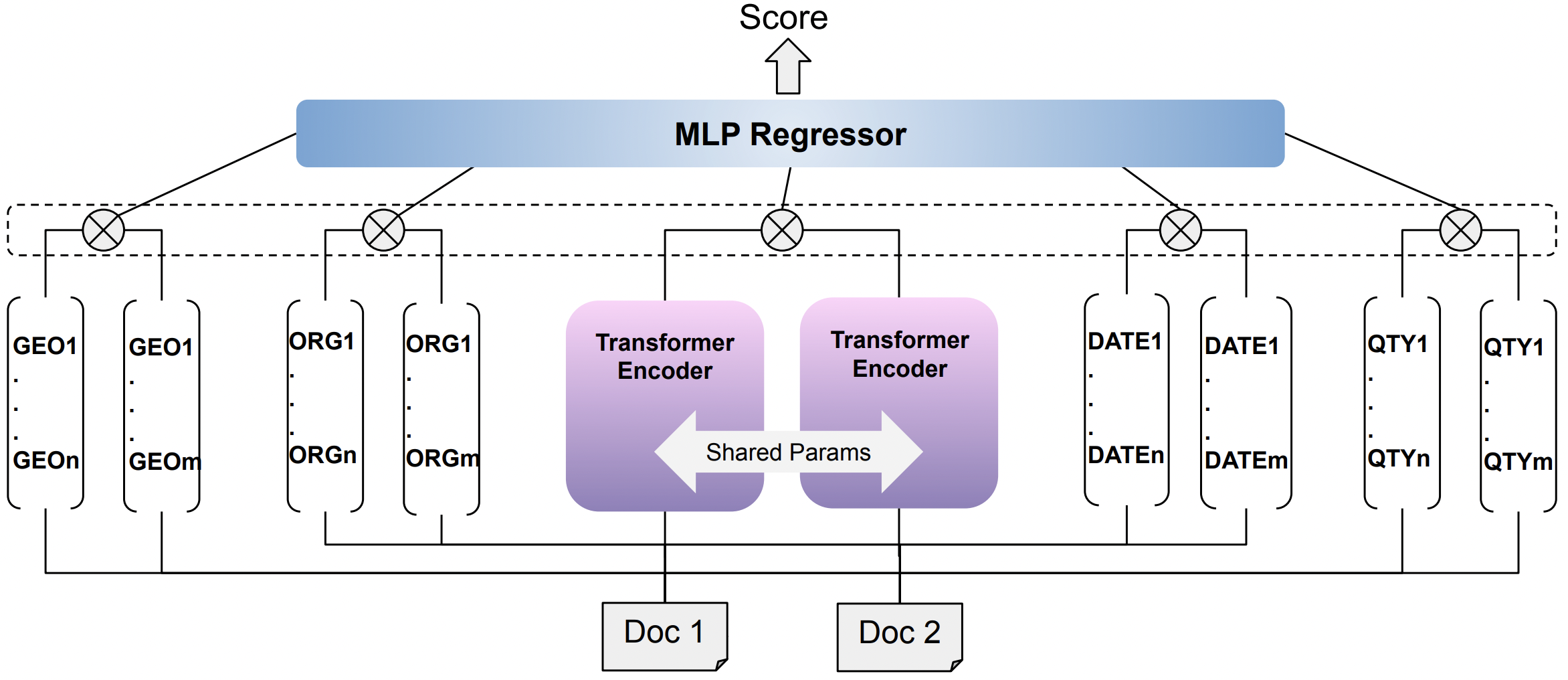}
    \caption{Entity-enriched Siamese Transformer model}
    \label{fig:model}
\end{figure*}

We propose an entity-enriched Siamese Transformer model which exploits multiple multilingual pre-trained Transformers (MPT) for multilingual news document similarity. The main sub-dimensions of similarity as annotated by the annotators are geolocation, time, shared entities and the shared narratives between the news articles \citep{taskpaper}\footnote{\url{https://competitions.codalab.org/competitions/33835}}. All this information is encoded in our model to capture different dimensions of news articles.
Figure \ref{fig:model} shows the architecture of our proposed model and section \ref{Model Details} discusses its details. 

\subsection{Model Details}
\label{Model Details}

\citet{reimers2019sentence} propose a Siamese and triplet network training methodology for the BERT-based \citep{devlin2018bert} models to derive semantically meaningful sentence embeddings. Instead of just learning the sentence representations, we use the Siamese (or bi-encoder) network architecture to learn document-level representations for the purpose of capturing the narrative of the article. 

In this work, we test multiple MPT models as the backbone Transformer \citep{vaswani2017attention} encoder for our Siamese network and train the model using a regression objective function \citep{reimers2019sentence}. In this, the model is trained such that it forces contextual representations from similar documents close to each other in contrast to dissimilar documents by reducing the mean squared error loss between the overall similarity score and the cosine similarity between document representations. 
First, we concatenate title and news article text together to generate a single document. For the documents that lie beyond the maximum input sequence length for the model, we truncate their length and use the initial part of the document as a proxy for its fundamental narrative that is usually stated in the title and the lead paragraphs of news articles.

In an effort to explicitly capture other dimensions of the news article such as mentions of geolocation, event date, organisations and other named entities, we use SpaCy NER\footnote{Pre-trained model en\_core\_web\_trf is used for our experiments. Ref. \url{https://spacy.io/models}} to extract entities from the documents. For the non-English documents, we use their machine translated English versions   
to extract the entities \citep{fan2021beyond}.\footnote{Pre-trained model m2m\_100\_418M is used for our experiments. Ref. \url{https://github.com/pytorch/fairseq}} All the extracted entities are aggregated into four different types of entities. The entity types along with their SpaCy labels (in brackets) are: 

\begin{itemize}
\itemsep0em 
    \item \textbf{Geolocation (GEO)}: Location (LOC) and geopolitical entities (GPE) which includes mentions of countries, cities, states etc.
    \item \textbf{Organization (ORG)}: Name of organization (ORG), person (PERSON) and mentions of other important named entities (FAC, EVENT, NORP, PRODUCT, WORK\_OF\_ART) which can include names of airports, highways, sports events, religious or political groups etc.
    \item \textbf{Date (DATE)}: Date of publish from training set and date (DATE) and time (TIME) extracted from the news articles.
    \item \textbf{Quantity (QTY)}: Mentions of numerical quantity (QUANTITY) which can include both ordinal and cardinal measurements (ORDINAL, CARDINAL).
\end{itemize}

After extracting the list of all the above mentioned entities for both the documents, we lowercase all the entities and compute the cosine similarity to get a single similarity score which we use as feature for further training. The complete algorithm to compute similarity between entities is presented at Algorithm \ref{alg1}.
\begin{algorithm}
\caption{Similarity between list of entities}
\begin{algorithmic} 
\Function{similarity}{$A, B$} 

\Comment{where A and B are list of entities}

\State Entities $\longleftarrow$ Union (A, B) 

\State Product $\longleftarrow$ Sum(A.count(ent)*B.count(ent) for ent in Entities)

\State MagnitudeA $\longleftarrow$ Sqrt(Sum(A.count(ent)$^{2}$ for ent in Entities)) 

\State MagnitudeB $\longleftarrow$ Sqrt(Sum(B.count(ent)$^{2}$ for ent in Entities)) 

\State Score $\longleftarrow$ Product / ( MagnitudeA * MagnitudeB )

\State return Score 
\EndFunction
\end{algorithmic}\label{alg1}
\end{algorithm}
We use the cosine scores of these four additional entity features along with the shared narrative score from Siamese Transformer to train a multilayer perceptron (MLP) with sigmoid activation to get the final news article similarity score. 
Overall, only the Siamese Transformer model is differentiable and all other are non-differentiable static features before the MLP layer. 
It is worth noting that both the Siamese Transformer and the MLP layer are trained separately and not jointly. The intuition behind using all these features together is to capture the similarity between news articles with different granularity and to assess the extent to which outlets write about same events.

\subsection{Semi-Supervised Learning}

One of the issues with the training dataset is its imbalanced nature, 
i.e. there are considerably more document pairs with low similarity scores as compared to pairs with high similarity scores (Figure \ref{fig:data_distri}). In order to upsample the instances with high similarity, we first augment the training set and then adopt a semi-supervised training methodology to train the model. For this, we first explore randomly sampling document pairs and label them using our previous best model trained on the training set. However, we found that random sampling does not generate good quality augmented training data as most of the generated pairs still lie in the similarity range of [0.0,0.2]. Therefore, we employ the augmentation strategy similar to one proposed by \citet{thakur-2020-AugSBERT}, i.e. BM25 sampling, for which we use the Elasticsearch\footnote{\url{https://www.elastic.co/elasticsearch/}} implementation of BM25 to generate augmented article pairs. However, our method is slightly different where we use the title of a news article as a query for retrieving other news article to get the document pairs. This helped the generation of documents pairs with similarity score that are more evenly distributed. A total of 59,943 additional document pairs were generated using the BM25 sampling and 2,845 document pairs were generated using machine translation. The complete details of the generation of augmented data is mentioned in Appendix \ref{appendix:augdata}. 

Semi-supervised learning \citep{zhu2009introduction} is a widely known training paradigm where a model is first trained on a human labelled dataset and the model is further used to extend the training set by automatically annotating the unlabelled dataset. Following previous studies \citep{thakur-2020-AugSBERT, jurkiewicz2020applicaai}, we initially start with training on the original training set and then for all the generated unlabelled document pairs, we use the previously trained model for inference to get the similarity scores for the new synthetic document pairs. Finally, we train our entity-enriched Siamese Transformer in a semi-supervised fashion on both the complete augmented training set. 

\section{Experimental Setup}
The training set 
contain 4,964 article pairs, however, after removing 
pairs with empty documents or if one of the document has less than ten tokens, we get a total of 4,544 article pairs for training. For the experiments, the training set is split into train and development sets using a 80:20 split ratio. The data is split in a stratified fashion to keep equal proportion of all languages, except Arabic document pairs that only appear in the development set, keeping it as 
a unseen language for the training set 
We get a total of 3,635 document pairs for the training set and 909 document pairs for the development set. In the end, we train models on the complete train and development set for final evaluation on the test set consisting of 4,953 document pairs. 

The training set 
has different types of labels which take into account different aspects of similarity among news articles (e.g., geolocation, narrative) \citep{taskpaper}. For our experiments, we use the ``Overall'' similarity as the true labels. However, the overall similarity label is in a range [1,4] where 1 signifies highest similarity and 4 means lowest similarity. In order to normalise these overall similarity label, we subtract all values from 4 to bring values in range [0,3], followed by min-max normalisation. The formula for this is as follows,
\begin{equation}
     x_{norm} = \frac{x - x_{min}}{x_{max} - x_{min}} = \frac{4 - x}{3}
\end{equation}
where $x$ denotes the overall similarity value and $x_{norm}$ is the normalised value in range [0,1] from least similar to most similar. We use these as labels for training our models.

Since there are no baselines provided by task organisers, we use strong baseline methods for comparison. First, we use separate entity-based cosine similarity scores (Section \ref{Model Details}) to find the Pearson correlation with the official test set labels. Second, we use Siamese multilingual BERT \citep{devlin2018bert}, XLM-RoBERTa base variant \citep{conneau2019unsupervised}, Universal Sentence Encoder (USE)\footnote{We employ a custom variant of USE model which supports 50+ languages. Ref. \url{https://huggingface.co/sentence-transformers/distiluse-base-multilingual-cased-v2}} \citep{yang2019multilingual} and LaBSE \citep{feng2020language}. Third, we also use the pre-trained MPNet\footnote{We employ a custom MPNet model which already fine-tuned on a 1 billion sentence pairs dataset with contrastive training objective. Ref. \url{https://huggingface.co/sentence-transformers/all-mpnet-base-v2}} \citep{song2020mpnet} and fine-tuned it on the machine translated English training set. Please refer to Appendix \ref{appendix:hyper} for training details and the hyperparameters used in our experiments.

\section{Results and Discussion}

Table \ref{tab:results} shows the Pearson correlation coefficient scores on the official test data. The first part of the Table \ref{tab:results} presents the results of the baseline methods and the second part presents the results of our proposed model. For baselines, we find that if we just consider the entity-based features without training, organisation (ORG) achieves the highest Pearson correlation with the official test set. Furthermore, for the trained Siamese Transformer models, the LaBSE works best and the XLM-RoBERTa achieves the lowest score of around 0.70. 

The second part of Table \ref{tab:results} shows the results of our entity-enriched Siamese Transformer trained on the augmented data using the semi-supervised learning paradigm. Training the Siamese Transformer on the augmented data brings significant improvements for all the models except USE when compared with the ones trained on the training set without augmentation. A potential explanation for LaBSE's impressive performance is the MLM and TLM pre-training objective on a dataset of 109 different languages, followed by training on a translation ranking task \citep{feng2020language}. We also test the Siamese Transformer both with and without the entity-enrichment in order to study how significant the improvements are statistically with entity features. Furthermore, we use Williams test \citep{graham2014testing} to test the statistical significance of increased correlation with the added auxiliary entity features. As shown in the Table \ref{tab:results}, for all Siamese Transformers trained on the augmented data, entity-enrichment brings improvements in results to a statistically significant degree (p-value$<$0.01). We also find that the results of entity-enriched Siamese Transformers are statistically significant (p-value$<$0.01) when compared with the baseline Siamese Transformers for all the models. Overall, our entity-enriched Siamese LaBSE model trained on augmented data achieves the highest Pearson correlation of 0.80164. 

\begin{table*}[!htbp]
\centering
\begin{tabular}{|l|c|c|}
\hline
\multicolumn{1}{|c|}{\textbf{Experiments}}                            & \textbf{Test Results} & \textbf{P-value}               \\ \hline 
\multicolumn{3}{|l|}{\textbf{Baselines}} \\
\hline
Geolocation (GEO)                                       & 0.26696      &                       \\
Organisation (ORG)                                      & 0.45503      &                       \\
Date (DATE)                                             & 0.43482      &                       \\
Quantity (QTY)                                          & 0.35791      &                       \\ 
Siamese mBERT                           & 0.73069      &                       \\
Siamese XLM-RoBERTa                                         & 0.70775      &                       \\
Siamese USE                                          & 0.73324      &                       \\
Siamese MPNet (English MT)                                          & 0.74536      &                       \\
Siamese LaBSE                                          & 0.79187      &                       \\
\hline 
\multicolumn{3}{|l|}{\textbf{Proposed Method}} \\
\hline

Siamese mBERT (Augmented data)                    & 0.76649      & \multicolumn{1}{l|}{} \\
Entity-enriched Siamese mBERT (Augmented data)    & 0.76771      & $<$0.0001               \\ \hline

Siamese XLM-RoBERTa (Augmented data)                    & 0.77669      & \multicolumn{1}{l|}{} \\
Entity-enriched Siamese XLM-RoBERTa (Augmented data)    & 0.77781      & $<$0.0001               \\ \hline

Siamese USE (Augmented data)                    & 0.73320      & \multicolumn{1}{l|}{} \\
Entity-enriched Siamese USE (Augmented data)    & 0.73590      & $<$0.0001               \\ \hline

Siamese LaBSE (Augmented data)                    & 0.80089      & \multicolumn{1}{l|}{} \\
Entity-enriched Siamese LaBSE (Augmented data)    & \textbf{0.80164}      & 0.0022               \\ \hline

\end{tabular}

\caption{Pearson correlation coefficient score on the official test data. The best performance is in bold.}
\label{tab:results}
\end{table*}

\begin{figure}[!htbp]
\includegraphics[width=\linewidth, height=4.5cm]{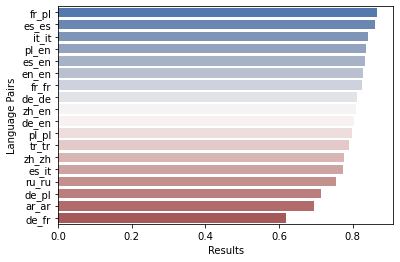}
\caption{Performance over different language pairs in official test data. Average score of mono-lingual and cross-lingual pairs is 0.79833 and 0.78205 respectively.}
\label{fig:results_per_lan}
\end{figure}

Figure \ref{fig:results_per_lan} presents the detailed analysis of the results for the entity-enriched Siamese LaBSE trained on augmented data model (best model). 
We observe that model performance varies in different language settings. The model performs the worst over German-French pairs (0.619), while it achieves the highest score on French-Polish articles (0.866). Overall, the model performs better on mono-lingual pairs than on cross-lingual pairs.  

We further analyse the "serious mistakes" that our best model tends to make: giving high/low similarity scores for dissimilar/similar article pairs. We identify these instances based on the difference between the true similarity score and the predicted similarity score from the model. We examine the cases when the absolute values of the differences are larger than 2.0 (total 46 samples found), and observe that: 1) News pairs that cover different stories around the same entities or topics are more challenging for the model. For example, the model outputs a similarity score of 1.77 for the following dissimilar news pair: an article about COVID-19 quarantine policy in Azerbaijan and another one regarding sanitary rules for preventing COVID-19 in Azerbaijan. 2) Web scraping could introduce noises in the model evaluation and obscure the true performance. We find that the scraping tool provided by the organizer fails to extract the real news content for all the Arabic news from Ahewar news website. The banner of the news website is returned instead. If we ignore the 25 pairs involving news from this website in the official test set, the Pearson correlation coefficient score over the Arabic data increases significantly from 0.69425 (2nd worst in Figure \ref{fig:results_per_lan}) to 0.84067 (4th best). 3) The model tends to overestimate the degree of similarity. Among the 46 "serious mistakes", only 15 of them are similar news pairs with predictions indicating dissimilarity. And these wrong predictions are all caused by scraping-related issues (the real content of news is not returned). However, we argue that our entity features may potentially increase the robustness and decrease the level of overestimation. We compare the performance between models with and without entity features. The results show that the differences between true label and prediction decrease for 22 of the 31 dissimilar news pairs. 

\section{Conclusion}
We introduce an entity-enriched Siamese Transformer for SemEval-2022 Task 8: Multilingual News Article Similarity. The error analysis shows that the entity-enrichment leads to statistically significant improvement and make models more robust for computing similarity between news articles in both monolingual and cross-lingual setting. Our entity-enriched LaBSE model achieves a Pearson Correlation of 0.802, ranking 2nd in the competition. The code is available at \url{https://github.com/iknoorjobs/semeval-code}.

\section{Acknowledgement}

This research has been partially supported by  a European Union – Horizon 2020 Program, grant no. 825297 (WeVerify), the European Union – Horizon 2020 Program under the scheme “INFRAIA-01-2018-2019 – Integrating Activities for Advanced Communities”, Grant Agreement n.871042 (“SoBigData++: European Integrated Infrastructure for Social Mining and Big Data Analytics” (http://www.sobigdata.eu)) and CDT in Speech and Language Technologies and their Applications funded by UKRI (grant no. EP/S023062/1).

\bibliographystyle{acl_natbib}

\bibliography{naacl2021.bib}

\appendix

\section{Appendix}

\subsection{Data Distribution}
\label{appendix:datadist}

\begin{figure}[h]
\includegraphics[width=\linewidth, height=5.5cm]{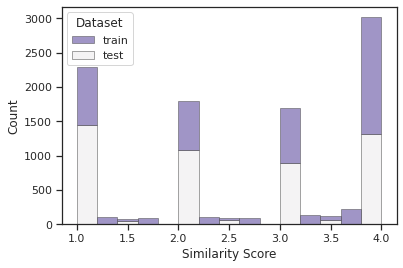}
\caption{Distribution of training and test data}
\label{fig:data_distri}
\end{figure}

\subsection{Augmented Data Generation}
\label{appendix:augdata}
The details of augmented data used in our experiments is as follows, 

\textbf{Augment Data 1.} In this, we used the training set provided by organisers where the title of the article is used as a query to all the other articles in the training set that are indexed in Elasticsearch. If title or article are not in English, then the English machine translated version is used to perform the retrieval. Also, we exclude the cases of document pairs that are already present in the training set. We generated 25,125 by using the top 5 retrieved documents for each case.

\textbf{Augment Data 2.} We machine translated the English document pairs with overall similarity score in range of [0.5-1] to other languages. We only choose to translate document pairs that are English to get a good quality machine translated augmented training data. For this we randomly sampled five different language pairs from the list of language pairs that follow the same distribution as that of language pairs found in the test set and machine translated the documents\footnote{Pre-trained model m2m\_100\_418M is used for our experiments. Ref. \url{https://github.com/pytorch/fairseq}} \citep{fan2021beyond}. We generated 2845 additional document pairs using this augmentation strategy. 

\textbf{Augment Data 3.} Here, we utilised ``All the news'' dataset available on Kaggle\footnote{\url{https://www.kaggle.com/snapcrack/all-the-news}} which contains 143,000 articles from 15 American publications. Here, title from the SemEval training set is used as query to all the articles in Kaggle new datasets. We generated 34,818 document pairs using the top 5 retrieved documents.

In our experiments, we use augment data 1, 2 and 3 which constitute a total of 62,788 additional document pairs and use them all together along with the training set to train the models.

\subsection{Hyperparameters}
\label{appendix:hyper}
The Transformer encoder of our Siamese Transformer (shared parameters) model is trained for 4 epochs with a batch size of 8, learning rate of 2e-5 and maximal input sequence length of 512. The output of model along with entity features is passed to the MLP layer which is composed of 32 hidden units trained with learning rate of 1e-3. The output of this MLP layer is used to get the final similarity score for the news articles. Adam \citep{kingma2014adam} is used to optimize the model parameters. For all the experiments, the model which works the best on the dev set is submitted and evaluated on the test set provided by the task organisers. All experiments are conducted on a machine with NVIDIA GeForce RTX 3090.

\end{document}